\documentclass[pdflatex,sn-mathphys]{sn-jnl}

\jyear{2021}%

\theoremstyle{thmstyleone}%
\theoremstyle{thmstyletwo}%

\theoremstyle{thmstylethree}%

\begin{document}

\title[ ]{A new visual quality metric for Evaluating the performance of multidimensional projections}


\author*[1,2]{\fnm{Maniru} \sur{Ibrahim}}\email{maniru.ibrahim@ul.ie}

\author[3]{\fnm{Thales} \sur{Vieira}}\email{thales@ic.ufal.br}


\affil*[1]{\orgdiv{Department of Mathematics \& Statistics}, \orgname{University of Limerick}, \orgaddress{\country{Ireland}}}

\affil[2]{\orgdiv{Department of Mathematics}, \orgname{COMSATS University Islamabad, Lahore Campus}, \orgaddress{\state{Lahore}, \country{Pakistan}}}

\affil[3]{\orgdiv{Institute of Computing}, \orgname{Federal University of Alagoas (UFAL)}, \orgaddress{ \state{Maceió}, \country{Brazil}}}


\abstract{Multidimensional projections (MP) are among the most essential approaches in the visual analysis of multidimensional data. It transforms multidimensional data into two-dimensional representations that may be shown as scatter plots while preserving their similarity with the original data. Human visual perception is frequently used to evaluate the quality of MP. In this work, we propose to study and improve on a well-known map called Local Affine Multidimensional Projection (LAMP), which takes a multidimensional instance and embeds it in Cartesian space via moving least squares deformation. We propose a new visual quality metric based on human perception. The new metric combines three previously used metrics: silhouette coefficient, neighborhood preservation, and silhouette ratio. We show that the proposed metric produces more precise results in analyzing the quality of MP than other previously used metrics. Finally, we describe an algorithm that attempts to overcome a limitation of the LAMP method which requires a similar scale for control points and their counterparts in the Cartesian space.}




\maketitle

\section{Introduction}\label{sec1}

Multidimensional data requires effective and interactive techniques to embed the data into a visual space. Multidimensional projections methods are among the most effective methods in the visual analysis of multidimensional data. Multidimensional projections have been used widely to transform multidimensional data into scatter plots, usually preserving similarity and the euclidean distances between the data points and their pattern \cite{1,2,11,17,21}.

Multidimensional projection has been successfully employed in several visualization applications such as pattern recognition, genetics \cite{13,15}, visual text analysis \cite{4,8}, word cloud analysis \cite{5,22}, vector field analysis \cite{6,18}, music and video \cite{11} to mention a few. Various surveys have been done on different multidimensional projections methods to compare their effectiveness, flexibility, computation efficiency, etc. \cite{3,7,10,12,16,20}.

Multidimensional projection aims to transform a discrete subset of high-dimensional space into a discrete subset of visual space preserving many features such as distance as much as possible. One of the most effective multidimensional projection techinques is LAMP \cite{11}, which is based on the Moving Least Squares (MLS) technique \cite{14,19}. In the MLS approach we associate with each point $p$ in our data set an affine mapping $f(p)$   and then the unique projection given by $f(p)=f_p (p)$. Joia et al. \cite{11} use moving least-squares minimization similar to that in rigid image deformation \cite{19} that uses subsets of the multidimensional data called the control points with their location in the visual space and use the control points to construct a collection of orthogonal mappings, one for each instance, and allow the user to modify the control points in the visual space to more promptly arrange them.

In this work, we propose to study the LAMP method \cite{11} and to improve its formulation, trying to overcome a limitation of the method that requires similar scaling between control points in the high dimensional data and their images in the visual space. We propose to learn a new visual quality metric for evaluating multidimensional projections, by solving an optimization problem. This metric combines a few metrics such as silhouette coefficient, neighborhood preservation and silhouette ratio between the original and projected data, which was used to evaluate LAMP projections. We use this novel learning algorithm to study the effect that scaling multidimensional data has on projection.

\section{Related Work}

As the quantity and complexity of visualization techniques have increased, choose which approach to use for any particular circumstance or application has gotten increasingly complicated. One example is multidimensional data visualization using projections, which has recently acquired popularity and is being used in an increasing variety of applications.

Multidimensional projections are vulnerable to errors and distortions. This is because orthogonal mappings from multidimensional spaces to visual spaces are only possible under some specific circumstances. To put it another way, neighborhood structures seen in the visual space may differ from those in the actual multidimensional space. Given that MDP-based visual analysis supposes that proximity relations in the visual space reflect similarities and that our visual perception is biased in favor of this assumption, the prospect of distortions between original and visual neighborhoods brings ambiguities that affect the human analytic process \cite{sacha}.

Most of the development of new approaches to evaluate projection quality was concerned with distance preservation and used assessment metrics that reflected that, such as stress and distance plots. The purpose, in many cases, is to give information for choosing or comparing projections. Graphical outputs of quality metrics are also employed, not only to aid in the discovery of optimal projections, but also to identify areas of a layout with superior characteristics or fewer deviations from target neighborhoods \cite{Mart}.

According to Nonato and Autepit \cite{errormetrics}, discrepancies in multidimensional projection mappings are primarily induced by two separate phenomena that impact visual space neighborhood structures. The first effect, known as Missing Neighbor (MN), occurs when neighboring instances in the multidimensional space are mapped far apart in the visual space. The second one is False Neighbor (FN). FN happens when non-neighboring instances in the multidimensional space are mapped near one another in the visual space. The operation of quantitative techniques to assess in multidimensional projection distortions varies significantly. Distinct measuring approaches attempt to quantify different components of the MN and FN phenomena, requiring global and/or local examinations and also techniques capable of determining specific distortion features. 

Various of these quality metrics were developed to support the evaluation and recommendation of projections. These quality metrics include  the correlation coefficient \cite{corr}, Kruskal's stress function \cite{stress},  silhouette coefficient \cite{tan}, neighborhood preservation \cite{11}, etc.

The correlation coefficient \cite{corr} attempts to quantify how distances in the original space are correlated to distances in the visual space. The correlation coefficient measures the difference between the distances between points in the original space graph and the distances between the points in the 2D embedding. Kruskal's stress function was proposed by Kruskal \cite{stress}, it measures how well original distances in the original high-dimensional space are preserved in the visual space.

The silhouette coefficient \cite{tan} assesses both the cohesiveness and the separation of clustered occurrences. The average distance between instance $x \in \mathcal{D}$ and all other instances in the same group as $x$ is used to compute $x's$ cohesiveness $a_x.$ The separation $b_x$ is the shortest distance between $x$ and all other instances of the same cluster. It is given by

$$Silh =\frac{1}{n}\sum\limits_{x \in \mathcal{D}}  \frac{(b_x - a_x)}{max(a_x, b_x)}.$$

The neighbourhood preservation metric \cite{11} computes the fraction of an instance's k-nearest neighbours who are still neighbours in the visual space.  A variant of the neighborhood preservation metric called Smooth Neighborhood Preservation (SNP) was introduced in \cite{paulo}, which takes into account both the number of neighbors retained in the projection and the distance that misplaced points are from their actual position. 

Other quality criteria, primarily to work with supervised MDP approaches, have been applied \cite{Grac}. Most present supervised multidimensional projections techniques do not account for human perceptual abilities, hence class structures may still be concealed to a human observer.  Several metrics have been introduced to model the human perception of class separability into supervised DR methods. These metrics include Distance  Consistency (DSC)  \cite{DSC}, Dunn’s index  \cite{Dunn},  LDA’s objective  \cite{Caco}, Silhouette coefficient. Sedlmair and Aupetit \cite{Sedl} created a machine learning framework to analyze all the measurements and discovered that DSC is the best one to understand how these metrics mimic human perception. They recently examined their suggested new metrics \cite{Aupe} and discovered that their new metrics (GONG) and (KNNG) perform much better than DSC. Peter extended KNNG and DSC with the ability to capture density information, so they can be properly used in an iterative  multidimensional projection process. But still, DSC and KNNG have been found to be among the best state-of-the-art measures but are still computationally efficient enough for our purpose. 

In this work, we propose a novel strategy by learning a quality metric for multidimensional projections.

\section{Proposed Method}
\subsection{Learning a New Measure to Evaluate Multidimensional Projections}\label{learnm}

Evaluating the quality of a projection is a very difficult task. For instance a projection may be good with respect to neighborhood preservation, while the same projection may be poor with respect to silhouette coefficient. To solve this problem, we define a new metric which combines three major measures for evaluating the quality of a projection. The new metric is a weighted sum of silhouette coefficient, neighborhood preservation, and silhouette ration. We find the weight of each metric by solving the weighted sum as an optimization problem. We create eighty (100) projections using three (3) different datasets (iris, wine, and vehicle datasets) and different scales. We give a grade from 1 (worse) to 5 (best) to each projection in an intuitive manner based on the layout of the projection. We use scikit-learn library in Python to split the data into training dataset $(70\%)$ and test dataset $(30\%).$ 

We solve the optimization problem as follows:

We first define the training dataset $X_{train} =\lbrace x_i\rbrace_{i=1}^{60}$ and its counterpart $Y_{train} = \{ y_i\}_{i=1}^{60},$ where $x_i = (m_1^{(i)}, m_2^{(i)}, m_3^{(i)}),$ $m_1$ is the silhouette coefficient, $m_2$ is the neighbourhood preservation, $m_3$ is the silhouette ratio, and $y_i$ is the grade of $x_i$. We need to learn the weight of a regression function $f: \mathbb{R}^3 \rightarrow \mathbb{R},$ given by $f_{\bar{w}}(m_1,m_2,m_3) = w_1m_1+w_2m_2+w_3m_3,$ where $\bar{w} = \{ w_1, w_2, w_3\}.$


We will minimize the loss
\begin{align}
 L(w_1,w_2,w_3) &= \sum\limits_{i=1}^{60} (f_{\bar{w}}(m_1^{(i)}, m_2^{(i)}, m_3^{(i)})-y_i)^2\\
 &=\sum\limits_{i=1}^{60}(w_1m_1^{(i)}+ w_2m_2^{(i)}+ w_3m_3^{(i)}-y_i)^2
\end{align}
with respect to the parameters $w_1,w_2,w_3.$ For the minimization to occur we need 
\begin{align}\label{min0}
    \frac{\partial L}{\partial w_1}=0, \;\; \frac{\partial L}{\partial w_2}=0 \;\; \frac{\partial L}{\partial w_3}=0.
\end{align}

Evaluating Equation \eqref{min0} and solving the partial derivative, we obtain
\begin{align*}
    0&=2\sum\limits_{i=1}^{60}(w_1m_1^{(i)}+ w_2m_2^{(i)}+ w_3m_3^{(i)}-y_i)m_1^{(i)}\\
    0&= 2\sum\limits_{i=1}^{60}(w_1m_1^{(i)}+ w_2m_2^{(i)}+ w_3m_3^{(i)}-y_i)m_2^{(i)}\\
    0& = 2\sum\limits_{i=1}^{60}(w_1m_1^{(i)}+ w_2m_2^{(i)}+ w_3m_3^{(i)}-y_i)m_3^{(i)}.
\end{align*}

Simplifying these equations to normal equations we get,

\begin{small}
\begin{equation*}
    w_1\sum\limits_{i=1}^{60}m_1^{(i)}m_3^{(i)}+w_2\sum\limits_{i=1}^{60}m_2^{(i)}m_3^{(i)}+w_3\sum\limits_{i=1}^{N}(m_3^{(i)})^2=\sum\limits_{i=1}^{60}y_im_1^{(i)}.
\end{equation*}
\end{small}

Substituting the values of $\{m_1^{i}, m_2^{i}, m_3^{i}\}_{i=1}^{60}$ into these Equation we obtain,
\begin{small}
\begin{align*}
    8.25391394w_1+5.41666345w_2+13.24087516w3&=67.4299\\
    5.41666345w_1+5.45949627w_2+14.00792903w_3&=51.3835\\
    13.24087516w1+14.00792903w_2+1095.0485935w_3&=116.8538.
\end{align*}
\end{small}

Solving the above system of linear equations using LU decomposition in Python we get,
\begin{align*}
    w_1&=5.7097, \;\; w_2 = 3.7741, \;\; and \;\; w_3 = -0.0106.
\end{align*}
Therefore, our new metric is
\begin{align*}
    M_{new}&=5.7097m_1+3.77416m_2-0.0106m_3.
\end{align*}

We use mean absolute error as our loss function to evaluate the model's performance. We then compute the mean absolute error on the test data. The in-sample error is the value of the loss function on the best fit model on the training set, while the out-of-sample error is the value of the loss function on the test set. 
\begin{align*}
MAE&=\frac{1}{N}\sum\limits_{i=1}^{N}\vert y_i - f_{\bar{w}}(x_i)\vert.
\end{align*}

\subsubsection{Hyperparameter Tuning Algorithm for Multidimensional Projections}

In this subsection, we describe a simple algorithm, where we sample an interval of different scales to find the best one, using the learned metric as the criterion. This is just an optimization problem, where we want to look for the scale that achieves maximum quality projection according to the new metric function.


\begin{algorithm}
\begin{algorithmic}[1]
    \State{Input: Dataset $D \subset \mathbb{R}^d$, Array $T,$ New metric, Scale $[a,b]$ uniformly sample}
    \For{each scale $s$ in $[a,b]$}
        \State{Scale the data with min = 0, and max = s}
        \State{Project the scaled data using Lamp}
        \State{Compute the new metric on the projection}
        \State{Store the value in the array $T$} 
    \EndFor
    \State{Find the highest value in the array $T$}
    \State{Find the scale corresponding to the position of the highest value in array $T$}
\end{algorithmic}
\caption{Hyperparameter tuning algorithm}
\label{alg:example2}
\end{algorithm}

\section{Results}
In this chapter, we perform an experiment based on the new metric we define. The results are split into four sections. Section \ref{result1} contains the results of the three measures ($m_1, m_2,$ and $m_3$) by projecting three datasets (iris, wine, and Vehicle Silhouettes datasets) with different scales using LAMP. Section \ref{result2} contains figures of some projections used to train the new metric, the values of $m_1, m_2,$ and $m_3$ for each projection, and also the grade is given to the projection intuitively based on its layout. In section \ref{result4}, we projected five different datasets (iris, wine, vehicle silhouettes, segmentation, and fish market datasets), and the show the results of the new metric for each. In section four, we evaluate the quality of the metric using cross-validation and also run Algorithm \ref{alg:example2} on the wine dataset to automatically find the best projection based on the learned metric and show the best figure.

\subsection{Datasets}
This section describes the datasets used in this study.

\begin{table}[h]
\centering
\begin{center}
\begin{minipage}{254pt}
\caption{Datasets}\label{tab1}%
\begin{tabular}{@{}lllll@{}}
\toprule
\textbf{Name} & \textbf{Size} & \textbf{Dim} & \textbf{Classes} & \textbf{Sources}\\
\midrule
Iris & 150 & 4 & 3 & \cite{Datasets} \\
Wine & 178 & 13 & 3 & \cite{Datasets} \\
Segmentation & 2,100 & 19 & 7 & \cite{Datasets}\\
Statlog (Vehicle Silhouettes) & 846 & 19 & 4 & \cite{Datasets} \\
Fish Market & 159 & 7 & 7 & \cite{Fish}\\
\botrule
\end{tabular}
\end{minipage}
\end{center}
\end{table}

\subsection{Evaluation of the influence of the scale parameter}\label{result1}
In this section, we compute the new metric on three different datasets (iris, wine, and vehicle silhouettes). In each dataset, we perform six projections each with a different scale (No scale, [0,0.01] x [0,0.01], [0,0.1] x [0,0.1], [0,1] x [0,1], [0,10] x [0,10], [0,100] x [0,100]) and also compute the scores of the metrics $m_1, m_2,$ and $m_3$ on each.

\begin{figure}[H]
        \centering
        \includegraphics[width=1\textwidth]{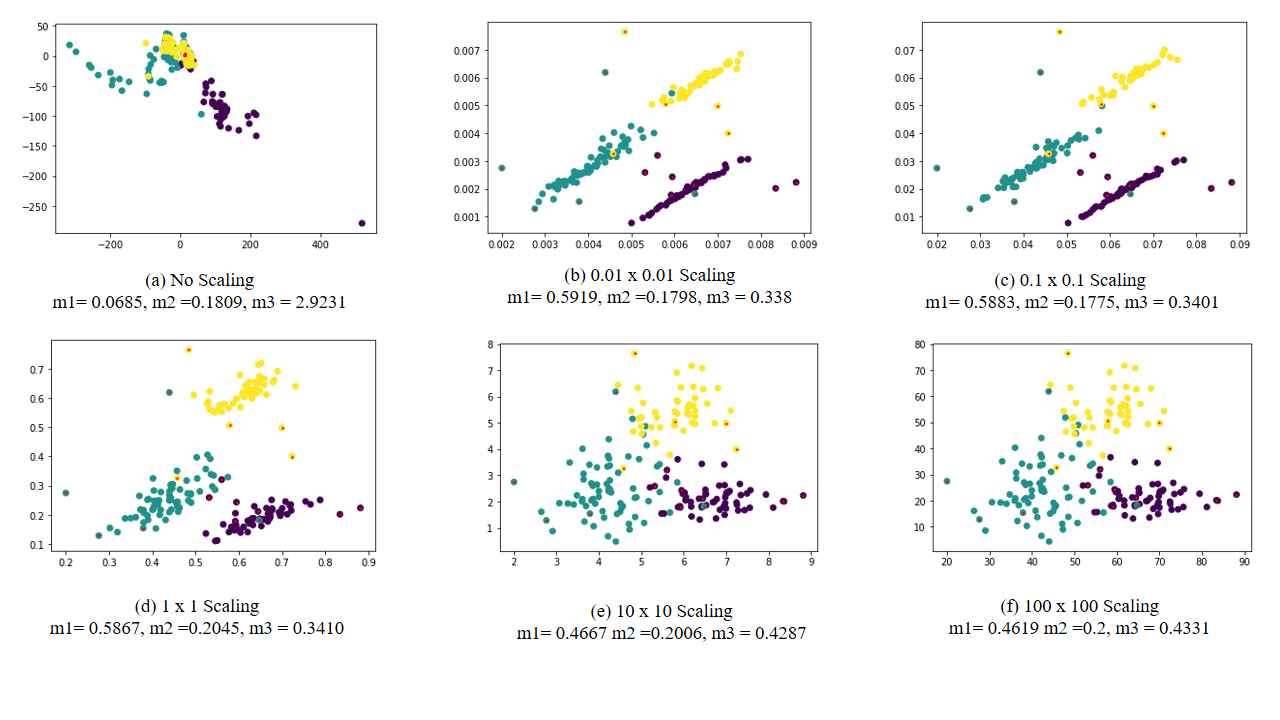}
      \caption{Figures showing projections on wine dataset with different scales.}\label{WEx}
\end{figure}

\begin{figure}[H]
        \centering
        \includegraphics[width=1\textwidth]{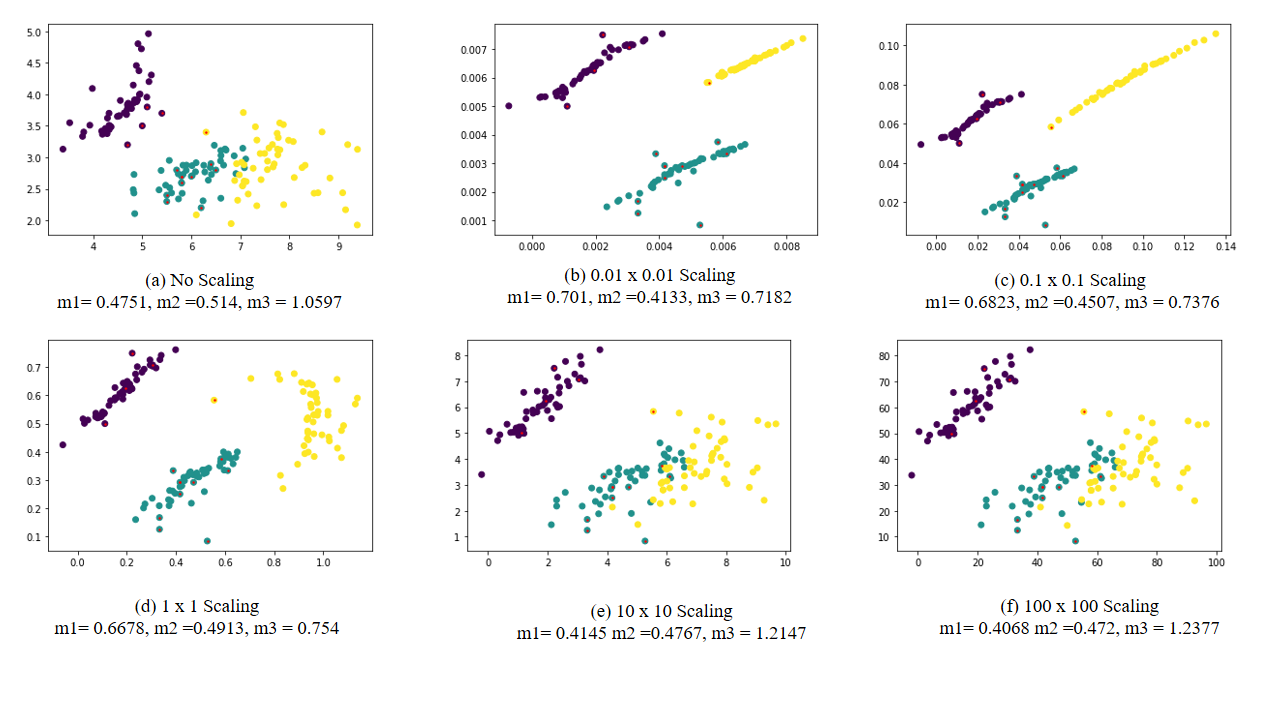}
      \caption{Figures showing projections on iris dataset with different scales.}\label{IEx}
\end{figure}

\begin{figure}[H]
        \centering
        \includegraphics[width=1\textwidth]{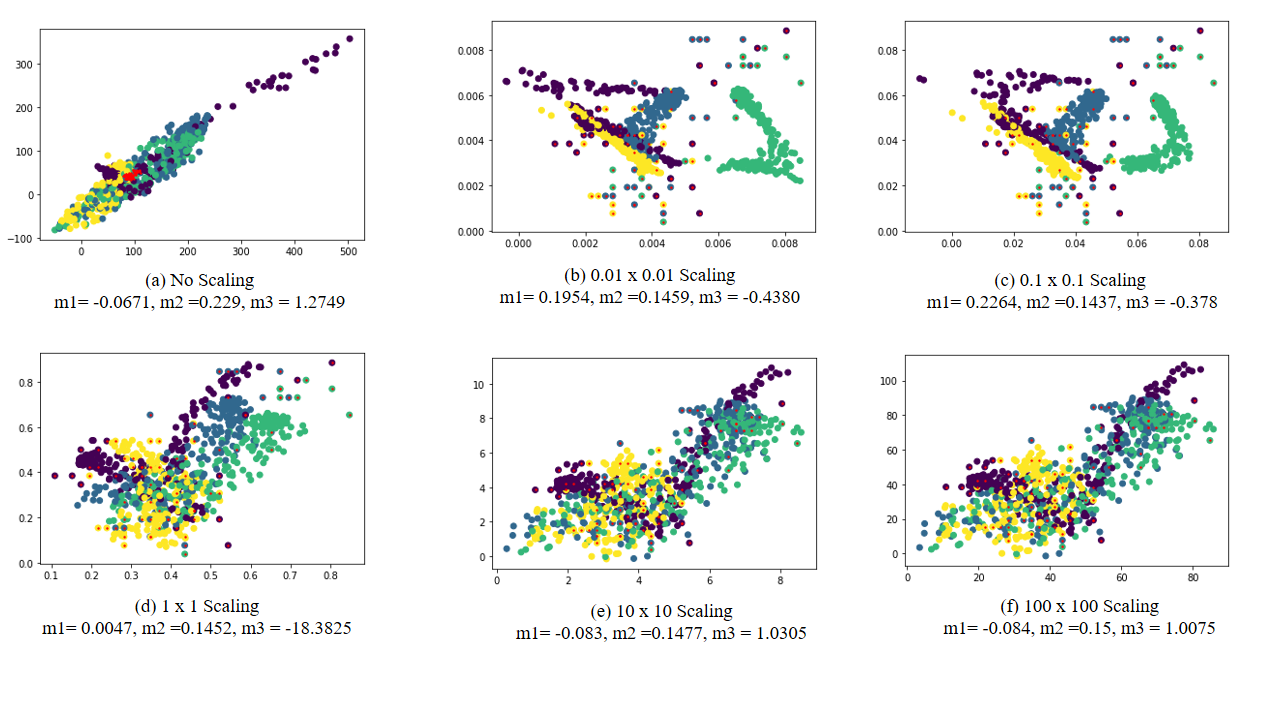}
      \caption{Figures showing projections on vehicle dataset with different scales.}\label{VEx}
\end{figure}

Figure \ref{WEx}, \ref{IEx}, \ref{VEx} show the result of the projection of wine, iris, and vehicle data respectively with different scales. The images show that depending on the scale, the quality of the resulting projection varies. This is very clear in the example in Figure \ref{IEx}(d) where the projection is good visually, and also with respect to $m_1$, $m_2,$ and $m_3$ metrics. Similarly, in Figure \ref{WEx}(d) the quality of the projection is visually good. Also, with respect to the $m_1$ and $m_3$ metrics, the projection is good, while with respect to the $m_2$ metric the projection is poor. This shows the importance of defining a new metric for assessing the quality of a projection by linearly combining the three metrics. Also, the result of the measures $m_1,$ $m_2,$ and $m_3$ was computed on the projections.

\subsection{Training a custom metric}\label{result2}

\begin{figure}[H]
        \centering
        \includegraphics[width=1\textwidth]{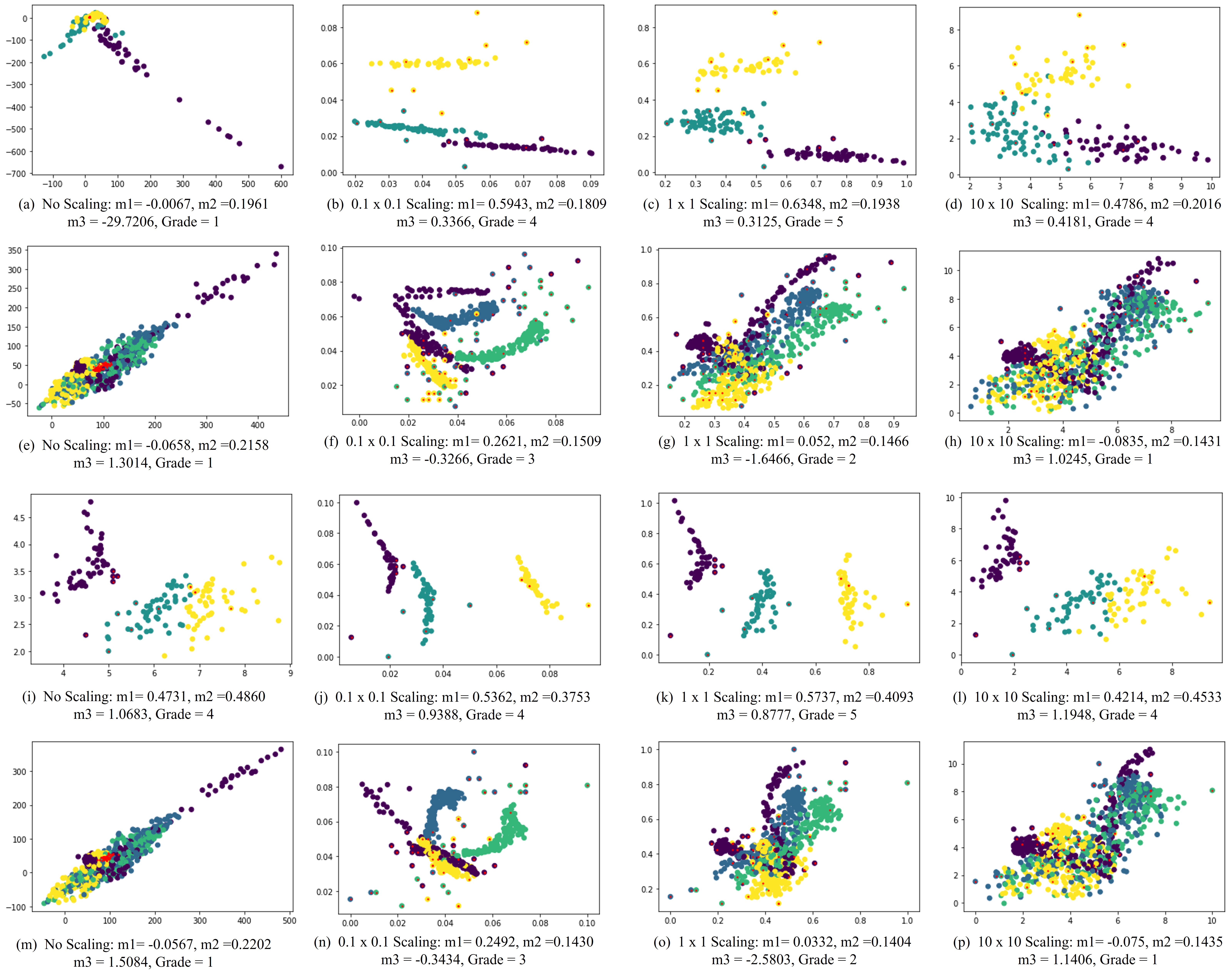}
      \caption{Figures showing projections used to train the new metric.}\label{figTrain}
\end{figure}

Figure \ref{figTrain} shows the image of the projections used to train the new metric and how each projection is labeled based on the quality of the projection for a specific scale. We also show the resulting value of $m_1, m_2,$ and $m_3$ for each projection. The metric training dataset comprises 60 examples from $X$ train datasets that have been manually labeled from $1$ to $5.$ The examples in Figure \ref{figTrain} demonstrate this. The visual quality of the projection is good in Figure 10(c), and the grade assigned to the new metric result is high. Figure \ref{figTrain}(h) shows the opposite, with a visually poor projection and a low grade was assigned to the new metric.

\begin{table}[h]
\centering
\begin{center}
\begin{minipage}{254pt}
\caption{The number of training and testing projections for each dataset}\label{tab1}%
\begin{tabular}{@{}llll@{}}
\toprule
\textbf{Dataset} & \textbf{Training} & \textbf{Test} & \textbf{Total}\\
\midrule
Iris & 21 & 7 & 28 \\
Wine & 21 & 7 & 28 \\
Statlog (Vehicle Silhouettes) & 18 & 6 & 24 \\
Total & 60 & 20 & 80\\
\botrule
\end{tabular}
\end{minipage}
\end{center}
\end{table}

\subsubsection{Visual Evaluation}\label{result3}
\begin{figure}[H]
        \centering
        \includegraphics[width=1\textwidth]{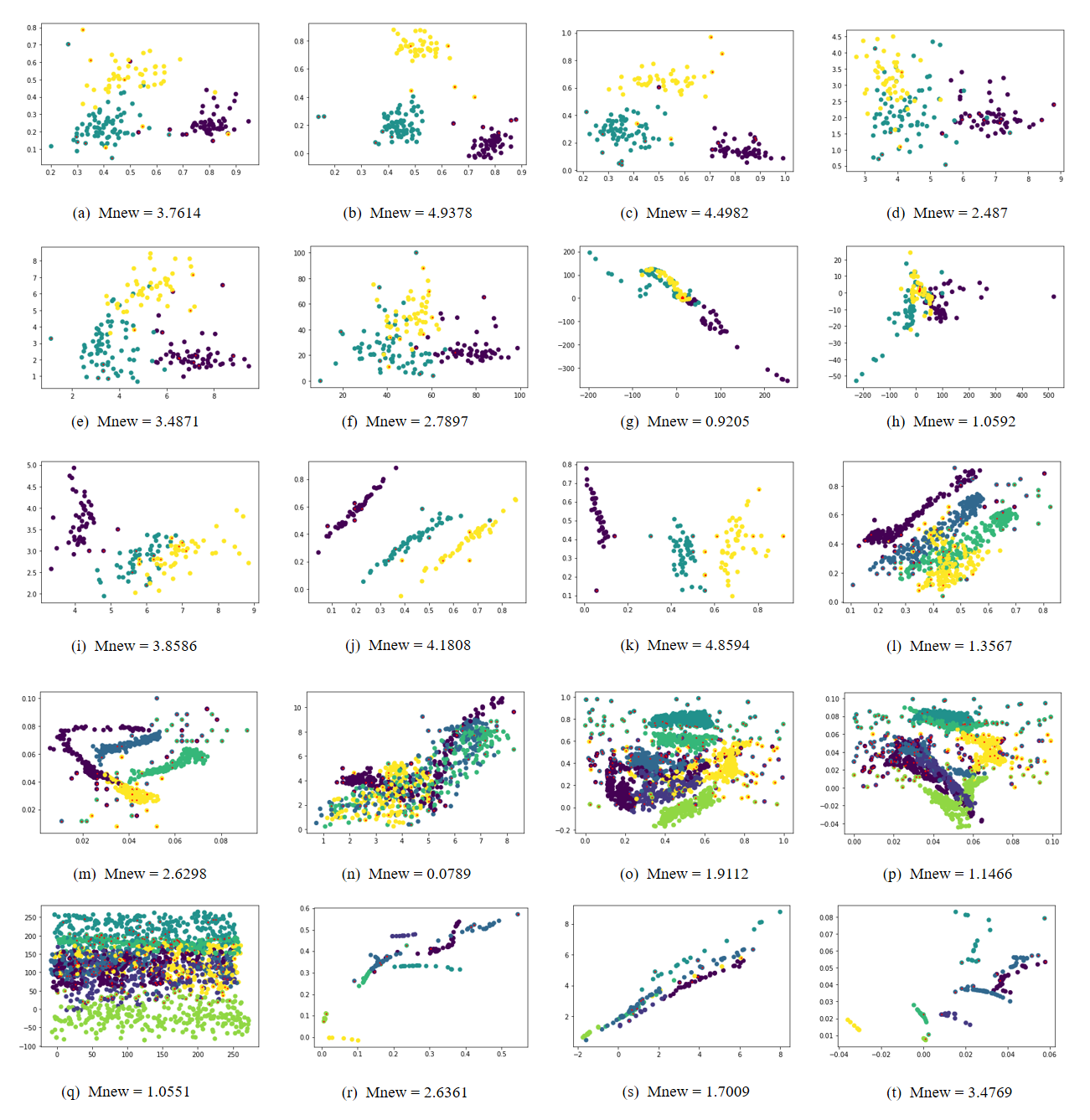}
      \caption{Figures showing projections with the result of the learnt metric.}\label{figVEv}
\end{figure}
To demonstrate how the learnt metric varies according to different situations. We perform twenty (20) projections using five (5) different datasets (iris, wine, vehicle silhouettes, segmentation, and fish market datasets). The images in Figure \ref{figVEv} show projections and the result of the learnt metric on the twenty (20) examples. 

In Figure \ref{figVEv}(b), the visual quality of the projection is good, and the new metric result is high. Figure \ref{figVEv}(g) depicts the opposite, with a poor visual projection and a low value for the new metric. In addition, the projection quality is moderate, and the new metric's value is moderate in Figure \ref{figVEv}(f).

\subsubsection{Cross-validation}\label{result4}
To evaluate how good our model is, we need to perform cross-validation on the test data by calculating the absolute error between each ground-truth label $x$ of the projection we gave and $y,$ which is the estimated quality computed using the learned metric. We then sum all the absolute errors and divide them by the total number of samples. The value we obtained is called the Mean Absolute Error (MAE), which is also our lost function. The value of MAE on the training dataset the out-sample error, while the value of MAE on the test dataset is the in-sample error.

We compute the mean absolute error $MAE$ on the train data containing $60$ samples as follows:
\begin{align*}
    MAE_{train}&=\frac{1}{n}\sum\limits_{i=1}^{n}\vert y_i - f_{\bar{w}}(x_i)\vert\\
    &=\frac{1}{60}\sum\limits_{i=1}^{60}\vert y_i - f_{\bar{w}}(x_i)\vert\\
    &=0.5414.
\end{align*}
Computing the median and standard deviation of the absolute errors, we get
\begin{align*}
    Median&=0.4398\\
    Std&=0.37
\end{align*}

We perform cross validation on the test data containing $20$ samples by computing the mean absolute error $MAE$
\begin{align*}
    MAE_{test}&=\frac{1}{n}\sum\limits_{i=1}^{n}\vert y_i - f_{\bar{w}}(x_i)\vert\\
    &=\frac{1}{20}\sum\limits_{i=1}^{20}\vert y_i - f_{\bar{w}}(x_i)\vert\\
    &=0.5630.
\end{align*}
Computing the median and standard deviation of the absolute errors, we get
\begin{align*}
    Median&=0.4893\\
    Std&=0.3878
\end{align*}

\begin{table}[h]
\centering
\begin{center}
\begin{minipage}{254pt}
\caption{The statistics of absolute errors of training and testing projections}\label{stats}%
\begin{tabular}{@{}llll@{}}
\toprule
\textbf{Dataset} & \textbf{MAE} & \textbf{Median} & \textbf{Standard Deviation}\\
\midrule
Training & 0.5414 & 0.4398 & 0.37 \\
Testing & 0.5630 & 0.4893 & 0.3878 \\
\botrule
\end{tabular}
\end{minipage}
\end{center}
\end{table}

The absolute error between the assigned and predicted values of the quality of the projection is between $0$ and $4.$ This is because the grades used to measure the quality of a projection range from $1$ to $5.$ 
Table 4 shows that the mean absolute error on the training data is $0.5414,$ which is less than $1,$ indicating that our model fits the data well. Similarly, the mean absolute error on the testing data is $0.5630,$ which is also less than $1,$ and very close to the MAE on the training data.  This also clearly shows that the model is good for prediction, implying that the new metric defined is very effective for determining the quality of a projection.

\begin{figure}
        \centering
        \includegraphics[width=.6\textwidth]{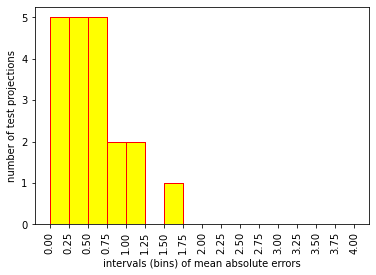}
      \caption{Histogram of error}\label{fighist}
\end{figure}

According to the histogram in Figure \ref{fighist}, out of the $20$ testing projections, $17$ projections $(85\%)$ have absolute errors in the interval $[0,1].$ Each of the intervals $[0,0.25],$ $[0.25,0.5],$ and $[0.5,0.75]$ accounts for $25\%$ of the total testing data, and $10\%$ of the data has absolute errors between $0.75$ and $1.$ Also, $2$ projections $(10\%)$ have absolute errors in the interval $[1,1.25],$ while $1$ projection $(5\%)$ have absolute errors in the interval $[1.50,1.75].$ This shows that the learnt metric have a less error, and so is good for determining the quality of a projection.

\subsubsection{Experimenting on Real-world Examples}

\begin{figure}[H]
        \centering
        \includegraphics[width=0.9\textwidth]{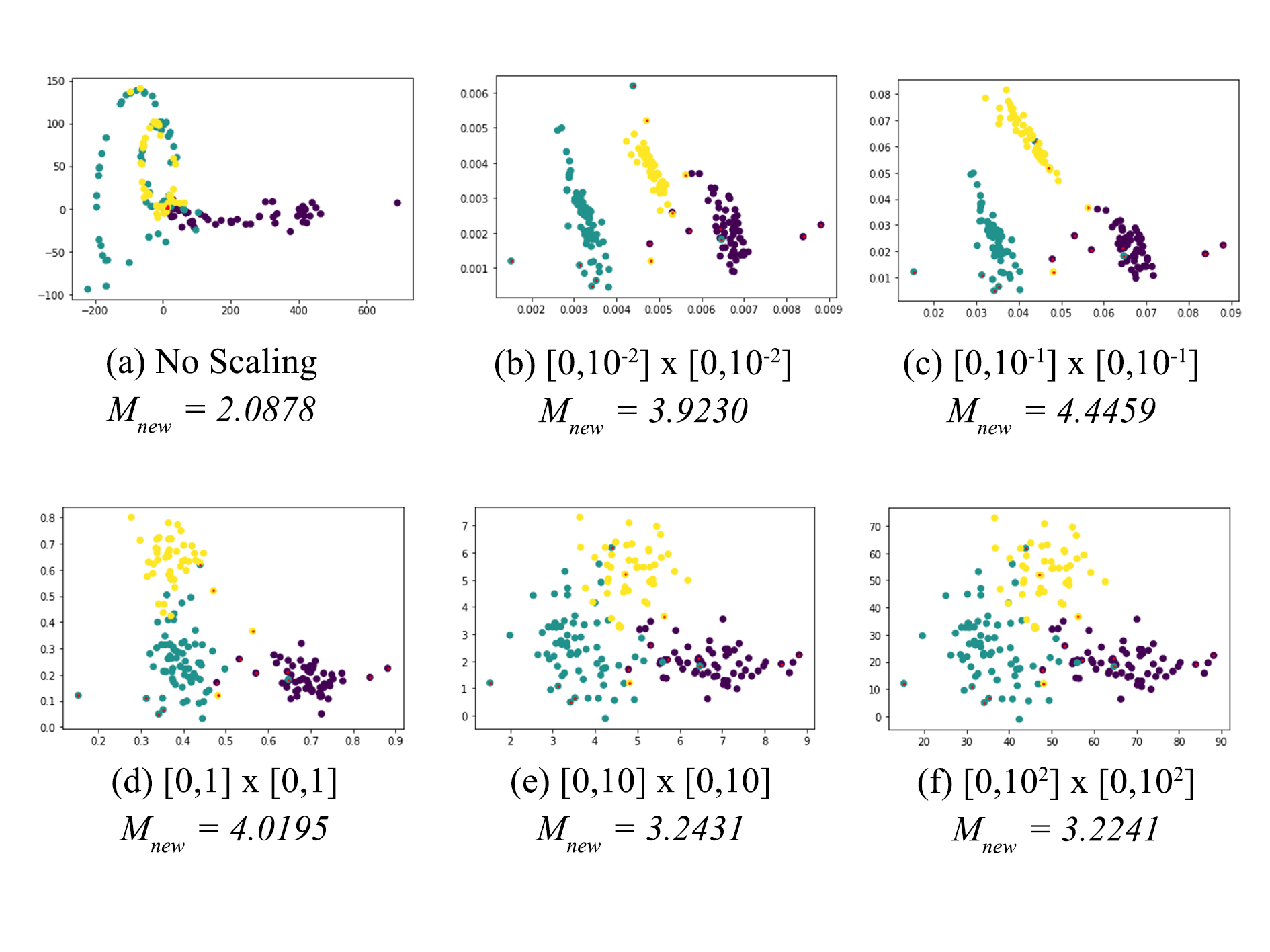}
      \caption{Running Algorithm \ref{alg:example2} on Wine dataset using interval of different scales}\label{figW1}
\end{figure}

Figure \ref{figW1} shows the results obtained by running Algorithm \ref{alg:example2} on Wine datasets using interval of different scales to automatically find the best projection. We empirically observed that the optimum scale is in the interval $[0.1, 1]$ in most cases. Thus, in the following example, We first employ the following scales. We run the algorithm on the scales ( $[0,10^{-2}] \times [0,10^{-2}],\;$ $[0,10^{-1}] \times [0,10^{-1}],\;$ $[0,1] \times [0,1],\;$ $[0,10] \times [0,10],\;$ \text{and} $\;[0,10^{2}] \times [0,10^{2}].$ (Figure \ref{figW1})

The scales $[0,0.1] \times [0,0.1]$ and $[0,1] \times [0,1]$ produce the best projection, as seen in the Figure \ref{figW1}. Therefore, in the next figure (Figure \ref{figW2}), the interval $[0.1, 1]$ was uniformly sampled into the scales ($[0,0.1] \times [0,0.1],\;$ $[0,0.2] \times [0,0.2],\;$ $[0,0.3] \times [0,0.3],\;$ $[0,0.4] \times [0,0.4],\;$ $[0,0.5] \times [0,0.5],\;$ $[0,0.6] \times [0,0.6],\;$ $[0,0.7] \times [0,0.7],\;$ $[0,0.8] \times [0,0.8],\;$ \text{and}  $\;[0,0.9] \times [0,0.9].$), and the optimum scale was found to be $[0,0.2] \times [0,0.2]$, as shown in Figure \ref{figW2} (b).

Figures \ref{figV1} and \ref{figV1} also indicate that when Algorithm \ref{alg:example2} is run on the Vehicle dataset, the optimal scale is $[0,0.2] \times [0,0.2].$

 \begin{figure}[H]
        \centering
        \includegraphics[width=0.9\textwidth]{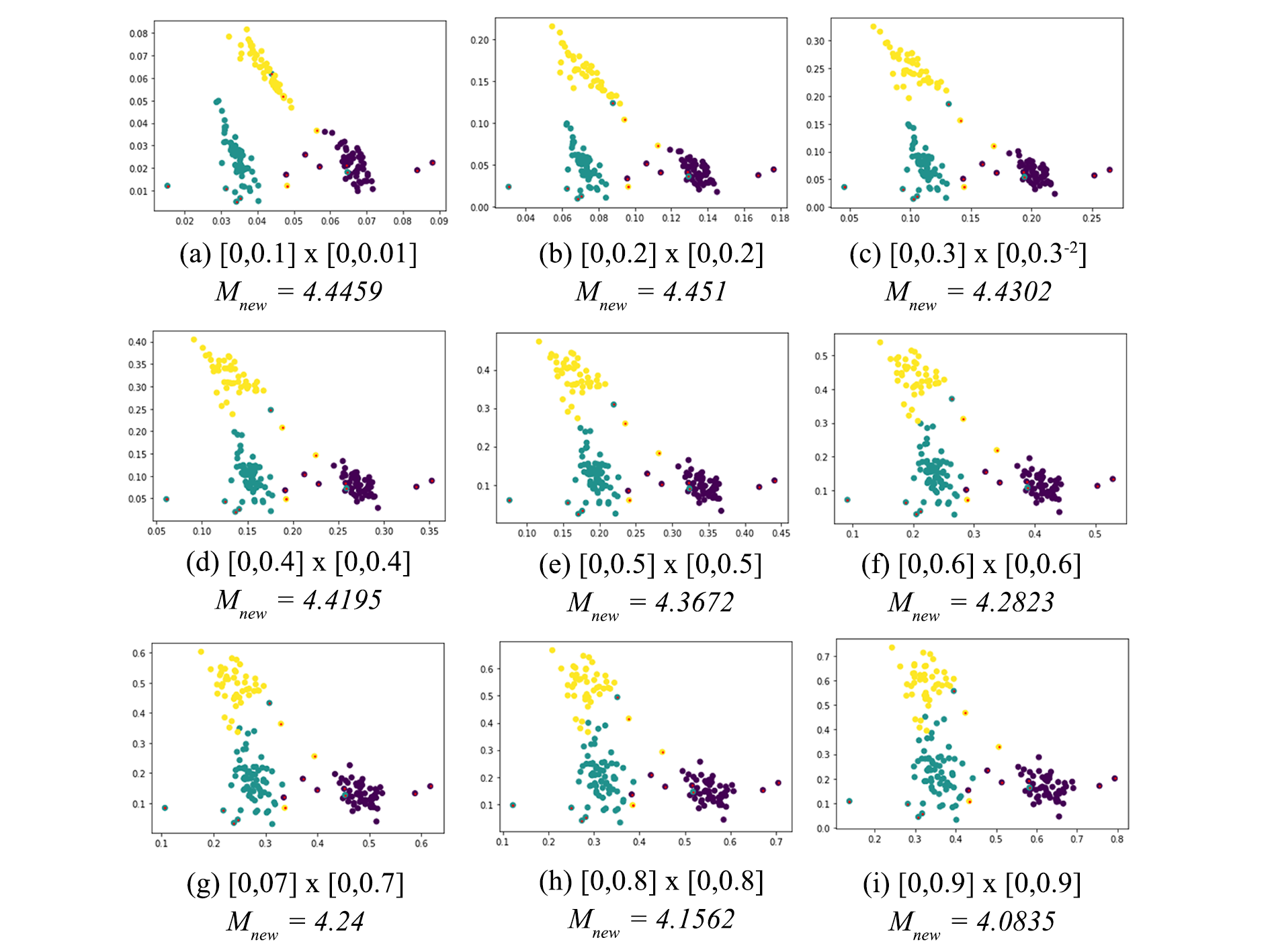}
      \caption{Running Algorithm \ref{alg:example2} on Wine dataset using interval of different scales}\label{figW2}
\end{figure}

  \begin{figure}[H]
        \centering
        \includegraphics[width=0.8\textwidth]{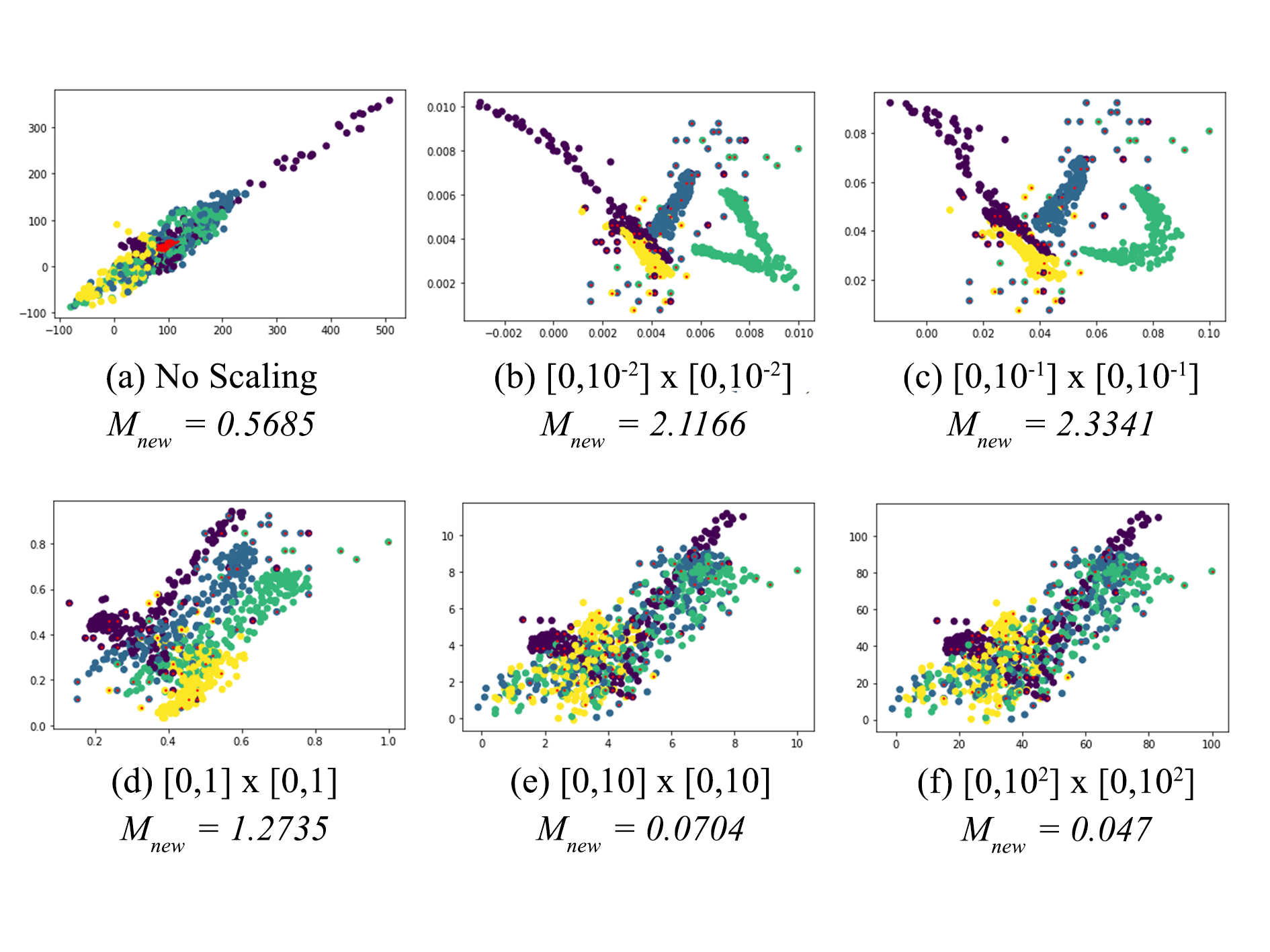}
      \caption{Running Algorithm \ref{alg:example2} on Vehicle Silhouettes dataset using interval of different scales}\label{figV1}
\end{figure}
 
\begin{figure}[H]
        \centering
        \includegraphics[width=0.9\textwidth]{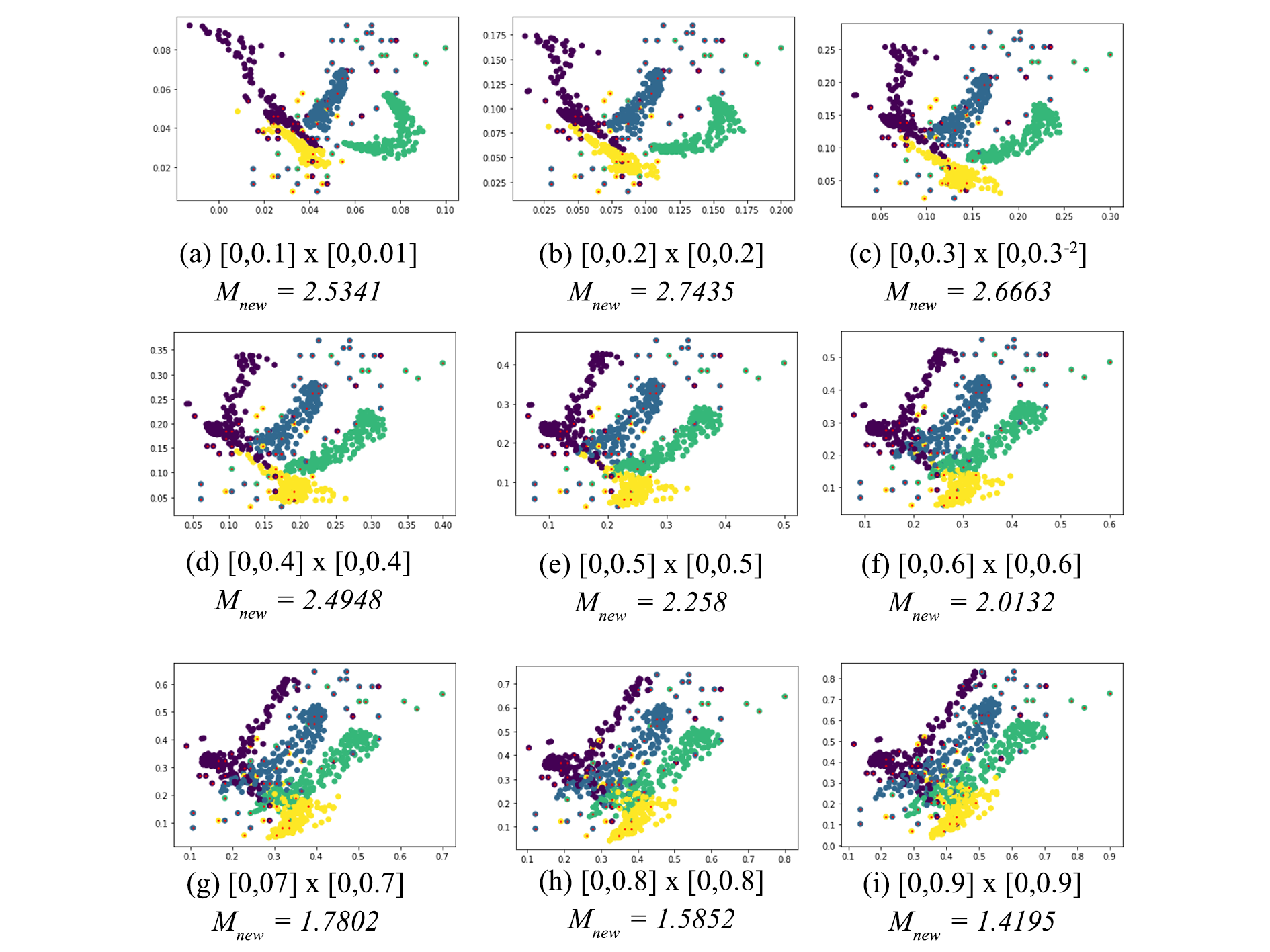}
      \caption{Running Algorithm \ref{alg:example2} on Vehicle Silhouettes dataset using interval of different scales}\label{figV2}
\end{figure}

\subsection{Discussion}
The results in Section \ref{result1} show that if we are given a fixed projection, the projection may be good with respect to one metric and bad with respect to another metric. For instance, in Figure \ref{WEx}(b) the projection is good with respect to silhouette coefficient, but it is bad with respect to neighborhood preservation. Therefore it is difficult to determine whether a projection is good since the quality of a projection is evaluated using the metrics. To solve this problem, we define a new metric $Mnew$ in Section \ref{learnm} by combining the three metrics. 

In Section \ref{result2}, we presented the results of the projection used to train the new metric. Figure 10 in Section \ref{learnm} illustrates the image of the projections used to train the new measure and how each projection is labeled dependent on the quality of the projection for a certain scale. 

According to the histogram in Figure \ref{fighist}, the absolute error of 17 samples (85 percent) is in the interval $[0,1],$ the absolute error of 3 samples (15 percent) is in the interval $[1,2],$ and there is no sample with absolute error in the interval $[2,5].$ This shows that the learned metric has a lower error rate and is good to be used in assessing the accuracy of a projection.

Figure \ref{figW1} and \ref{figW2} depicts the results produced by executing Algorithm \ref{alg:example2} on Wine datasets to identify the optimal projection using intervals of different scales. According to the algorithm, Figure \ref{figW2} (b) with the scale $[0,0.2] \times [0,0.2]$ provides the optimal projection. Similarly, the algorithm was computed on a different dataset (Vehicle data), and the optimal scale was found to be $[0,0.2]  \times [0,0.2]$ (Figure 
\ref{figV2} (b)).

\section{Conclusion}

We developed a new metric for evaluating the impact of scale on the quality of a projection in this paper. In several scenarios, the proposed metric has been found to be very effective. We also show that the scales of the multidimensional dataset have an impact on the quality of the projection. As a result, we built an algorithm for determining the scales that produce the best projection for every given dataset. It was empirically observed that the optimal scale that gives the best projection lies in the interval $[0.1,1].$ Another element that need to be look into more is determining the optimal number of neighbours to generate the desired layout.
A radius of impact to each control point might be defined as an alternative to the $k-$nearest neighbours technique used in our approach. 

\bmhead{Acknowledgments}
Author (Maniru Ibrahim) gratefully acknowledge acknowledge the financial support of the Science Foundation Ireland (SFI) under Grant Number SFI 18/CRT/6049 and the International Centre for Theoretical Physics (ICTP) Grant Number AF-18/19-01.

\section*{Declarations}

\subsection*{Ethical approval}
Not Applicable
\subsection*{Availability of supporting data}
Not Applicable
\subsection*{Competing interests}
The Authors declare that there is no conflict of interest.

\subsection*{Funding}
This work has emanated from research conducted with the financial support of the Science Foundation Ireland (SFI) under Grant Number SFI 18/CRT/6049 and the International Centre for Theoretical Physics (ICTP) Grant Number AF-18/19-01.  

\subsection*{Authors' contributions}
Maniru Ibrahim wrote the full manuscript. Thales Vieira reviewed and supervised the research. All authors read and approved the final manuscript. 

\subsection*{Acknowledgments}
Author (Maniru Ibrahim) gratefully acknowledge acknowledge the financial support of the Science Foundation Ireland (SFI) under Grant Number SFI 18/CRT/6049 and the International Centre for Theoretical Physics (ICTP) Grant Number AF-18/19-01.


\bibliography{sn-bibliography}


\end{document}